\documentclass[journal]{IEEEtran}

\usepackage{amsmath, amssymb, epsfig, color}
\usepackage{booktabs}
 \usepackage{graphicx}
\usepackage{subcaption}
\usepackage{url}
\usepackage{xurl}

\usepackage{multirow, array, booktabs}

\newcommand{\PreserveBackslash}[1]{\let\temp=\\#1\let\\=\temp}
\newcolumntype{C}[1]{>{\PreserveBackslash\centering}p{#1}}
\newcolumntype{R}[1]{>{\PreserveBackslash\raggedleft}p{#1}}
\newcolumntype{L}[1]{>{\PreserveBackslash\raggedright}p{#1}}

\let\OLDthebibliography\thebibliography
\renewcommand\thebibliography[1]{
  \OLDthebibliography{#1}
  \setlength{\parskip}{0pt}
  \setlength{\itemsep}{0pt plus 0.3ex}
}

\begin{document} 
\sloppy

\title{Fast Attention-Based Simplification of LiDAR Point Clouds for Object Detection and Classification}

\author{
    Zoltan Rozsa,~\IEEEmembership{Member,~IEEE},
    Ákos Madaras,
    Qi Wei,
    Xin Lu,~\IEEEmembership{Senior Member,~IEEE},
    Marcell Golarits,
    Hui Yuan,~\IEEEmembership{Senior Member,~IEEE},
    Tamas Sziranyi,~\IEEEmembership{Life Member,~IEEE},
    and Raouf Hamzaoui,~\IEEEmembership{Senior Member,~IEEE}%
    \thanks{
        Z. Rozsa, Á. Madaras, M. Golarits, and T. Sziranyi are with the Institute for Computer Science and Control (SZTAKI), Budapest, Hungary.
    }%
    \thanks{
        Q. Wei is with the School of Information Engineering, Nanchang University, Nanchang, China.
    }%
    \thanks{
        X. Lu and R. Hamzaoui are with the Faculty of Technology, Arts, and Culture, De Montfort University, Leicester, U.K.
    }%
    \thanks{
        H. Yuan is with the School of Control Science and Engineering, Shandong University, Jinan, China.
    }%
    \thanks{
        (Corresponding author: Raouf Hamzaoui.)
    }%
}

\maketitle


%
\begin{abstract}
LiDAR point clouds are widely used in autonomous driving and consist of large numbers of 3D points captured at high frequency to represent surrounding objects such as vehicles, pedestrians, and traffic signs. While this dense data enables accurate perception, it also increases computational cost and power consumption, which can limit real-time deployment. Existing point cloud sampling methods typically face a trade-off: very fast approaches tend to reduce accuracy, while more accurate methods are computationally expensive. To address this limitation, we propose an efficient learned point cloud simplification method for LiDAR data. The method combines a feature embedding module with an attention-based sampling module to prioritize task-relevant regions and is trained end-to-end. We evaluate the method against farthest point sampling (FPS) and random sampling (RS) on 3D object detection on the KITTI dataset and on object classification across four datasets. The method was consistently faster than FPS and achieved similar, and in some settings better, accuracy, with the largest gains under aggressive downsampling. It was slower than RS, but it typically preserved accuracy more reliably at high sampling ratios.
\end{abstract}


%
\section{Introduction}
\label{sec:intro}

Light detection and ranging (LiDAR) sensors have emerged as one of the most promising technologies for autonomous driving. LiDAR sensors are used for object detection as an alternative to or in conjunction with video cameras and radar. Using time of flight to measure the distance between a vehicle and objects, LiDAR sensors can help create an accurate representation of the surroundings of the vehicle in the form of three-dimensional point clouds. However, the huge amount of data forming these point clouds poses a significant challenge for processing, storage, and transmission.

Due to real-time and embedded constraints, processing all points is often impractical. Thus, point cloud downsampling is commonly applied before subsequent tasks.

FPS, which is one of the traditional downsampling approaches, only considers the Euclidean distance between points. Although it can cover the whole point cloud uniformly through iterations, it does not consider the semantic representation of the point cloud, which may limit the performance of subsequent downstream tasks.

Building on the success of deep learning-based methods for point cloud recognition \cite{2} \cite{1} \cite{5} \cite{6} and generation tasks \cite{8} \cite{17} \cite{18}, researchers developed deep learning-based downsampling methods to simplify the point cloud while aiming to preserve task-relevant semantic features. Although these methods can retain semantic information for downstream tasks, downstream performance can decline at high sampling ratios.

The geometric structure of the sampled point cloud can differ from that of the original one. For example, sampling may concentrate around important points and neglect other parts of the original point cloud. If only semantic features are prioritized, the geometric structure may not be preserved well, while focusing only on geometric structure may not guarantee identification accuracy. Therefore, a key problem in point cloud sampling is to balance semantic features and geometric structure.

In \cite{CAS}, we proposed a learning-based sampling method called CAS-Net. The objective of this network is to optimize a downstream task while preserving the geometric structure of the point cloud. Given an input point cloud, CAS-Net first extracts point-wise features using a feature embedding module that applies a grouping layer to collect $k$ neighbours per point, followed by a multi-layer perceptron. To better preserve global geometric information, the input point cloud is duplicated $k$ times and combined with the grouped features. CAS-Net then refines these features with an attention-based sampling module composed of stacked offset-attention (OA) layers. Finally, it estimates a learnable sampling matrix and applies it to the input point cloud to produce a simplified representation. CAS-Net was shown \cite{CAS} to outperform several previous learning-based simplification methods \cite{3} \cite{4} \cite{12} \cite{13} \cite{16} on the ModelNet40 benchmark for point cloud classification when PointNet \cite{1} is used as the downstream classifier.

In this paper, our goal is to adapt CAS-Net for LiDAR-based object detection and classification in real-world autonomous driving scenarios. A key challenge is reducing the computational complexity of the original model, which is currently too high for real-time deployment.

Our main contributions are as follows:

\begin{itemize}
\item We validate CAS-Net for 3D object detection on the KITTI dataset using PointPillars \cite{Lang_2019_CVPR} and show that it preserves detection performance under aggressive downsampling.
\item We show that, for 3D object detection on KITTI, CAS-Net outperforms RS and FPS at high downsampling ratios, while requiring less downsampling time than FPS.
\item We evaluate CAS-Net for 3D object classification on four datasets (ModelNet40, KITTI, ScanObjectNN, and ESTATE).
\item We show that reducing both the neighborhood size $k$ and the number of OA layers to one substantially reduces runtime, with small performance changes in stable settings and less predictable changes on noisier data.
\item We compare three neighborhood search implementations (PyTorch3D ball query, PyTorch3D brute-force k-NN, and a CPU-based k-d tree) to analyse the trade-off between speed and accuracy.
\end{itemize}

The remainder of this paper is organized as follows. In Section II, we briefly review related work on point cloud sampling. In Section III, we present the proposed sampling network. Experimental results and conclusions are given in Sections IV and V, respectively.

\section{Related Work}

Point cloud simplification methods can be divided into traditional methods and learning-based methods. 

\subsection{Traditional methods}

Traditional methods are task-agnostic and use fixed geometric or statistical rules to downsample point clouds. Common examples include RS, FPS, and Poisson-disk sampling \cite{22}. In RS, points are selected with uniform probability, which can miss fine structures or rare regions, especially at high downsampling ratios. FPS is an iterative procedure that repeatedly selects the point that maximizes its distance to the already selected set under a Euclidean metric, producing well-spread samples but with higher computational cost on large point clouds. Poisson-disk sampling enforces a minimum distance between sampled points, yielding evenly distributed samples; its runtime depends on the implementation and often requires acceleration structures for efficiency.

\subsection{Learning-based methods}

Dovrat \emph{et al.} \cite{3} proposed S-Net, a generative sampling network based on the PointNet architecture \cite{1}. The network generates a smaller set of points optimized for a downstream task, which are then matched to their nearest neighbors in the input point cloud to obtain a valid subset. However, this nearest-neighbor matching step occurs as a post-processing stage and is not part of the trainable model. SampleNet \cite{4} addressed this limitation by introducing a differentiable relaxation of point cloud sampling, where sampled points are approximated as mixtures of input points during training. However, the method may still struggle to preserve fine local details. Wang \emph{et al.} \cite{13} proposed PST-Net, which uses a transformer-based self-attention mechanism to model geometric relationships among points and learn a task-specific resampling distribution for point cloud downsampling. The method is permutation-invariant and robust to noise. Lin \emph{et al.} \cite{12} proposed DA-Net, which addresses density variation using a density-adaptive k-nearest neighbor module and improves robustness to noise through a local adjustment module. The method is trained end-to-end to produce task-oriented sampled points for point cloud classification. However, similar to several learning-based sampling approaches, the generated points are not guaranteed to be a strict subset of the input point cloud, which may affect geometric consistency with the original shape. Nezhadarya \emph{et al.} \cite{11} proposed a Critical Points Layer (CPL), which learns to reduce the number of points while preserving those that are most important for the task. CPL can be combined with graph-based point cloud convolution layers to form a neural network architecture. Qian \emph{et al.} \cite{10} proposed MOPS-Net, a deep learning method for task-oriented point cloud downsampling based on a matrix optimization formulation. The method models downsampling as a discrete point selection problem and relaxes the binary constraints to obtain a differentiable optimization problem. A neural network is then designed to approximate this optimization by capturing both local and global structures of the input point cloud. Yang \emph{et al.} \cite{14} introduced point attention transformers for point cloud processing. The model uses a parameter-efficient group shuffle attention mechanism to replace standard multi-head attention, enabling efficient learning of relationships between points while maintaining permutation equivariance and handling inputs of varying size. The method also proposes Gumbel subset sampling (GSS), an end-to-end learnable and task-agnostic sampling operation that selects representative subsets of points. GSS uses the Gumbel–Softmax technique to produce soft subsets during training and discrete subsets during inference. Sun \emph{et al.} \cite{16} proposed straight sampling, an end-to-end discrete sampling method that outputs a hard subset of points during training. The approach learns sampling alternatives from semantic features and uses a straight-through estimator to enable gradient propagation through the discrete sampling operation within a hierarchical point cloud learning framework. Guo et al. \cite{Guo25} formulated point sampling as a Top-k selection problem and made it trainable end-to-end using a differentiable Top-k approximation based on entropy-regularized optimal transport. 
Li et al. \cite{Li25} formulated point sampling as a maximum a posteriori (MAP) inference problem, where the sampling objective is guided by upsampling and perceptual reconstruction. The method uses residual graph convolution networks and a pseudo-residual link to exploit prior information. 
Dehghanpour et al. \cite{Dehghanpour25} proposed a point-based transformer for point cloud downsampling that models relationships among points while preserving permutation invariance. The resulting transformer-based sampling module is task-oriented and robust to noise, and it improves downstream segmentation by selecting points that better preserve fine geometric details.
Wu et al. \cite{Wu25} proposed a method that learns shape-specific sampling policies using a sparse attention map that integrates global and local cues. The method then applies bin-based learning to control the number of points selected from different score ranges, which enables a controllable trade-off between preserving sharp local details and maintaining global sampling uniformity.
Sugimoto et al. \cite{Sugimoto25} proposed a SampleNet-based downsampling method that combines average voxel grid sampling, which divides the point cloud into a voxel grid and uses the average coordinates within each voxel as representative points, with a modified FPS to increase point density in critical regions. The method aims to balance the preservation of overall object shape with the retention of fine local details for 3D object classification.

Compared with traditional methods, learned-based methods tend to trade improved task performance for higher computational cost, which becomes more pronounced as the input point cloud grows.

\begin{figure*}[t]
\centerline{\includegraphics [scale=0.15]{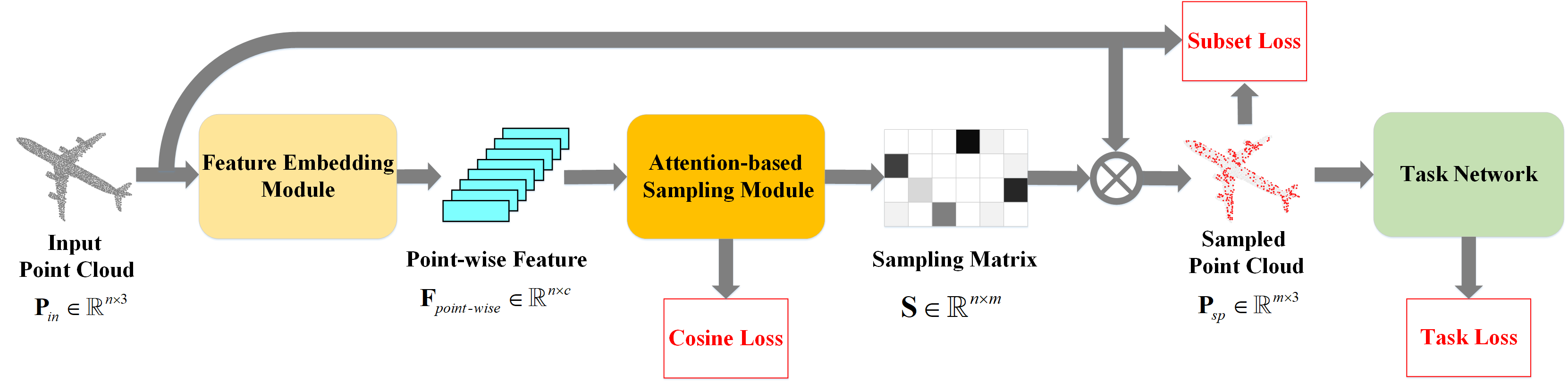}}
\caption{CAS-Net Architecture.}
\label{fig:arch}
\end{figure*}

\section{CAS-Net}
\label{sec:intro2}

The goal of CAS-Net is to find an optimal solution for a specific downstream task while maintaining the geometric structure of the point cloud. The architecture of CAS-Net is shown in Fig.~\ref{fig:arch}. Given an input point cloud $\mathbf{P}_{\mathrm{in}} \in \mathbb{R}^{n \times 3}$ and a network for a specified task, CAS-Net first uses a feature embedding module to capture local and global features.

Based on the resulting feature map, an attention-based sampling module enhances informative features and estimates a sampling matrix \textbf{S}. The downsampled point cloud \textbf{P$_{sp}$} is then obtained by multiplying \textbf{S} and \textbf{P$_{in}$}. Finally, \textbf{P$_{sp}$} is fed into the task network. The entire procedure is trained end-to-end using a composite loss function.

\subsection{Feature embedding module}
As shown in Fig.~\ref{fig:asm}, the input to the feature embedding module is an unordered point cloud with \emph{n} points. First, the unordered points are encoded by mapping the input to a vector in a higher-dimensional space.

To extract point-wise features from the input point cloud, we adopt the grouping layer \cite{2} due to its high efficiency. For a given point \textbf{p} in the input point set \textbf{P$_{in}$}, the grouping operation is defined as
\setlength{\abovedisplayskip}{2.2pt}
\setlength{\belowdisplayskip}{2.2pt}
\begin{eqnarray}
Group(\textbf {p})=
\{
\textbf {p$_1$}- \textbf {p},
\textbf {p$_2$}- \textbf {p}
,...,
\textbf {p$_k$}- \textbf {p}
\},
\end{eqnarray}
where \textbf {p$_i$}, \emph{i=}1,...,\emph{k}, represent the \emph{k} neighboring points of \textbf{p}. The output of the grouping layer can be written as
\setlength{\abovedisplayskip}{2pt}
\setlength{\belowdisplayskip}{2pt}
\begin{eqnarray}
\textbf {F$_{group}$}=Group
\left(\textbf {P$_{in}$}\right).
\end{eqnarray}

To better preserve global geometric information, the input point cloud is duplicated \emph{k} times and concatenated with the grouped features:
\setlength{\abovedisplayskip}{2.2pt}
\setlength{\belowdisplayskip}{2.2pt}
\begin{eqnarray}
\textbf {F$_{combine}$}=concat \left(
\underbrace{\textbf {P$_{in}$},\textbf {P$_{in}$}
,...,
\textbf {P$_{in}$}}_{k\ times},\textbf {F$_{group}$}
\right).
\end{eqnarray}

Finally, a multi-layer perceptron (MLP) $\sigma( \cdot )$ maps the combined feature to 
\setlength{\abovedisplayskip}{2pt}
\setlength{\belowdisplayskip}{2pt}
\begin{eqnarray}
\textbf {F$_{pointwise}$}=\sigma(
\textbf {F$_{combine}$}), 
\end{eqnarray}
where $\mathbf{F}_{\mathrm{pointwise}} \in \mathbb{R}^{n \times c}$. 

\subsection{Attention-based sampling module}
This module aims at selecting points from existing points. This is done with an attention mechanism that identifies and captures the most informative points during training. The self-attention (SA) mechanism is defined as
\begin{eqnarray}
\mathbf{Q} = \mathbf{F}_{in}\mathbf{W}_{q},\quad
\mathbf{K} = \mathbf{F}_{in}\mathbf{W}_{k},\quad
\mathbf{V} = \mathbf{F}_{in}\mathbf{W}_{v},
\end{eqnarray}
\begin{eqnarray}
\mathbf{F}_{sa}
= \text{softmax}\left(
\frac{\mathbf{Q}\mathbf{K}^{T}}{\sqrt{d_{k}}}
\right)\mathbf{V},
\end{eqnarray}
where $\mathbf{F}_{in}$ represents the input features,  
$\mathbf{W}_{q}$, $\mathbf{W}_{k}$, and $\mathbf{W}_{v}$ are learnable linear projections with the same output dimension, and $d_{k}$ is the dimensionality of $\mathbf{K}$.  
The self-attention layer is computed as
\begin{eqnarray}
\mathbf{F}_{out}
= \gamma(\mathbf{F}_{sa}) + \mathbf{F}_{in},
\end{eqnarray}
where $\gamma(\cdot)$ denotes an MLP. However, the self-attention layer cannot handle the problem of information loss when the network is deeper \cite{20}. By taking the difference between the attention features and the input features into account, an offset attention (OA) \cite{15} is used to modify the feature
\setlength{\abovedisplayskip}{2pt}
\setlength{\belowdisplayskip}{2pt}
\begin{eqnarray}
\textbf {F$_{out}$}=OA
\left(
\textbf {F$_{in}$}
\right)
= \gamma
\left(
\textbf {F$_{in}$}-\textbf {F$_{sa}$}
\right)+\textbf {F$_{in}$}.
\end{eqnarray}

\begin{figure}[t]
\centerline{\includegraphics[scale=0.15]{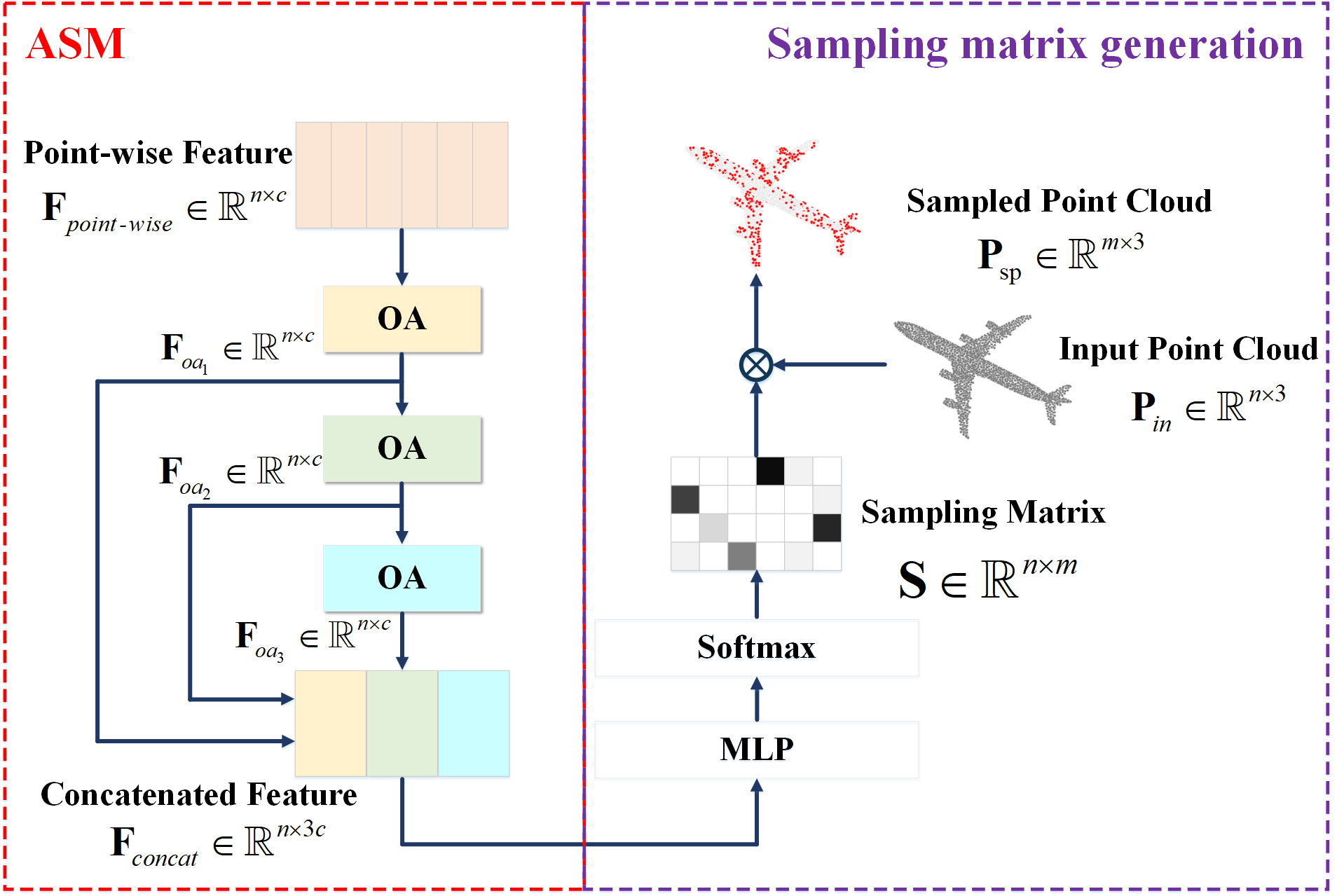}}
\caption{Overview of ASM.}
\label{fig:asm}
\end{figure}

To preserve the geometric features together with the semantic features, information fusion is needed between layers. Therefore, an attention-based sampling module (ASM) consisting of three skip-connected OA layers (Fig.~\ref{fig:asm}) is used. The output of each layer is then concatenated along the feature dimension:

\setlength{\abovedisplayskip}{2pt}
\setlength{\belowdisplayskip}{2pt}
\begin{eqnarray}
\mathbf{F}_{oa_1} = OA(\mathbf{F}_{pointwise}), \\
\mathbf{F}_{oa_2} = OA(\mathbf{F}_{oa_1}), \\
\mathbf{F}_{oa_3} = OA(\mathbf{F}_{oa_2}).
\end{eqnarray}

\setlength{\abovedisplayskip}{2pt}
\setlength{\belowdisplayskip}{2pt}
\begin{equation}
\mathbf{F}_{concat}
= \text{concat}(\mathbf{F}_{oa_1},\, \mathbf{F}_{oa_2},\, \mathbf{F}_{oa_3}), 
\end{equation}
where $\mathbf{F}_{oa_i}$ represents the output feature of the \emph{i}-th OA layer and $\mathbf{F}_{concat}$ is the concatenated feature.

\subsection{Sampling matrix generation}
The concatenated feature ${\mathbf{F}_{concat}}$ is then used to predict a learnable soft sampling matrix $\tilde{\mathbf{S}}$ through an MLP followed by a softmax function:
\begin{eqnarray}
\tilde{\mathbf{S}}
= \mathrm{softmax}\!\left( \rho(\mathbf{F}_{concat}) \right),
\label{soft}
\end{eqnarray}
where $\rho(\cdot)$ denotes the MLP. To obtain a binary sampling matrix $\mathbf{S}$, the largest element of each
column of $\tilde{\mathbf{S}}$ is set to one and the remaining elements to zero. This strategy ensures
that the sampled points form a subset of the input point cloud.

Given a sampling matrix, the input point cloud can be downsampled via matrix multiplication.
However, since the hard sampling matrix $\mathbf{S}$ is not differentiable, its
gradient is approximated during backpropagation using the gradient of $\tilde{\mathbf{S}}$, following the
straight-through estimator \cite{16}. Accordingly, two variants are proposed: the attention-based hard sampling network (AHSN) and the attention-based soft
sampling network (ASSN). The only difference between them is that AHSN uses the hard
sampling matrix $\mathbf{S}$ in the forward pass, while ASSN uses the soft sampling matrix
$\tilde{\mathbf{S}}$ directly. The downsampled point cloud produced by ASSN is computed as
\begin{eqnarray}
\mathbf{P}_{sp} = \tilde{\mathbf{S}}^{T} \mathbf{P}_{in},
\end{eqnarray}
whereas the downsampled point cloud for AHSN is obtained using
\begin{eqnarray}
\mathbf{P}_{sp} = \mathbf{S}^{T} \mathbf{P}_{in}.
\end{eqnarray}

\subsection{Loss function}

The proposed network is trained using a joint loss function that consists of three components:
\begin{eqnarray}
\begin{aligned}
L_{total}=&L_{task}(\textbf{P$_{sp}$})+
\alpha L_{subset}(\textbf{P$_{in}$},\textbf{P$_{sp}$})\\&+
\beta L_{cosine}
(\tilde{\textbf{S}}),
\end{aligned}
\end{eqnarray}
\setlength{\abovedisplayskip}{0.2pt}where $L_{task}(\cdot)$ encourages the network to learn an optimized downsampled point set for the downstream task, $L_{subset}(\cdot)$ preserves the geometry structure of the downsampled point cloud, and $L_{cosine}(\cdot)$ is used to reduce the number of duplicate points. The parameters $\alpha$ and $\beta$ are weights that balance the contribution of each component. Specifically,  $L_{subset}(\cdot)$ ensures that \textbf{P$_{in}$} and \textbf{P$_{sp}$} are close to each other as follows \cite{3}\cite{4}: 
\begin{eqnarray}
\begin{aligned}
L_{subset}(\textbf{P$_{in}$},\textbf{P$_{sp}$})=&\frac{1}{|\textbf{P$_{in}$}|}\sum_{x \in \textbf{P$_{in}$}}^{} \min_{y \in \textbf{P$_{sp}$}}||x-y||_2^2 
\\&+\frac{1}{|\textbf{P$_{sp}$}|}\sum_{y \in \textbf{P$_{sp}$}}^{} \min_{x \in \textbf{P$_{in}$}}||y-x||_2^2. 
\end{aligned}
\end{eqnarray}
The first term ensures that every point in \textbf{P$_{in}$} has a nearby point in \textbf{P$_{sp}$}, while the second term ensures that the points in \textbf{P$_{sp}$} are distributed all over \textbf{P$_{in}$}. However, the network may focus on a few important points, leading to an accumulation of duplicated points in the downsampled point set \cite{12}. To address this issue, the cosine loss $L_{cosine}(\cdot)$ \cite{16} is introduced:
\begin{align}
L_{cosine}(\tilde{\mathbf{S}}) =
\sum_{i \neq j}
\left| \cos(\tilde{\mathbf{s}}_i, \tilde{\mathbf{s}}_j) \right|,
\end{align}
where $\tilde{\mathbf{s}}_i$ and $\tilde{\mathbf{s}}_j$ denote row vectors of $\tilde{\mathbf{S}}$ (\ref{soft}).

\section{Experimental Results}

\subsection{Object detection using LiDAR point clouds}

The following experiments are motivated by the large data volumes in intelligent transportation systems, where efficient subsampling is essential for real-time processing and efficient data storage. Similar requirements are also important in other perception sensors, such as automotive radar \cite{10315142}.  This task therefore provides a suitable framework to validate our algorithm's effectiveness in a practical, high-demand scenario. The experiments were conducted on an NVIDIA A100 GPU with 40 GB of memory.

In these experiments, in the original CAS-Net code \footnote{\url{https://github.com/yuanhui0325/CAS-Net}} the ball query implementation was replaced by PyTorch3D ball query and the PointNet classification network was substituted with the PointPillars detection network \cite{Lang_2019_CVPR}\footnote{\url{https://github.com/madak88/CAS-Net_Fast-Attention-Based-Simplification-of-LiDAR-Point-Clouds-for-Object-Detection}}. We selected PointPillars for its strong balance between inference speed and accuracy, making it a suitable candidate to evaluate the real-time applicability of our sampling method. Although newer models such as ViKIENet \cite{Yu_2025_CVPR} improve accuracy and approaches that use tracking feedback continue to appear \cite{10147030}, PointPillars remains widely used in real-time embedded systems. This is important because accurate object segmentation \cite{10848132} and detection \cite{8818349} are essential for autonomous driving and enable safe navigation through reliable perception. 

We used the labeled part of the KITTI 3D Object Detection benchmark for all experiments. We followed the standard training/validation split and report results on the validation set. Also, following standard practice, only points within the camera's Field of View (FoV) (which are annotated) were retained.

For fair comparison with previous work, we cropped each input point cloud to a maximum of 8192 points (all resulting segments of the original dataset were included in the evaluation). This limit served two purposes:
    \begin{enumerate}
\item To reduce the runtime of FPS. As FPS has a worst-case complexity of $\mathcal{O}(N^2)$, we split each point cloud of size $N$ into $M$ smaller chunks and applied FPS independently within each chunk. This yields an approximate $M$-fold reduction in total runtime without affecting detection performance as shown in recent work \cite{9919246}.
\item To keep CAS-Net within practical GPU memory limits. CAS-Net stores $k$ neighbors per input point. The memory required grows with the number of input points. We therefore cropped the input at 8192 points.
    \end{enumerate}

In Table \ref{tab:comparison_det}, we report the Moderate Mean Average Precision ($\text{mAP}_{\text{mod}}$), which is the standard metric used on the KITTI benchmark as performance measure. In CAS-Net (AHSN), we used ball query for neighborhood search in the feature embedding module, with $k=32$ and a radius of $2$. We used $c=64$ features per point and set $\alpha=1$ and $\beta=1$ in the loss function. The network was trained for 160 epochs with a batch size of 8 and a learning rate of $2.5 \times 10^{-4}$.

\begin{table}[h]
\centering
\small
\setlength{\tabcolsep}{3pt} 
\caption{Comparison of 3D Object Detection Performance (Moderate mAP \%) and downsampling time (s) on the KITTI validation set.}
\label{tab:comparison_det}
\begin{tabular}{l c cc cc cc}
\toprule
Method & $D=1$ & \multicolumn{2}{c}{$D=2$} & \multicolumn{2}{c}{$D=4$} & \multicolumn{2}{c}{$D=8$} \\
\cmidrule(lr){2-2} \cmidrule(lr){3-4} \cmidrule(lr){5-6} \cmidrule(lr){7-8}
& mAP & mAP & Time & mAP & Time & mAP & Time \\
\midrule
RS      & 64.46 & 53.12 & 0.001 & 37.45 & 0.001 & 22.22 & 0.001 \\
FPS     & 64.46 & \textbf{62.24} & 0.144 & 49.85 & 0.075 & 20.94 & 0.041 \\
CAS-Net & 64.46 & 61.79 & 0.072 & \textbf{56.74} & 0.038 & \textbf{47.97} & 0.029 \\
\bottomrule
\end{tabular}
\end{table}

All methods started from the same reference performance at $D=1$, but CAS-Net’s advantage increased as the downsampling ratio grew. At $D=8$, CAS-Net achieved a Moderate mAP of 47.97\%, outperforming RS (22.22\%) and FPS (20.94\%), which indicates better preservation of task-relevant geometric structure under aggressive simplification. CAS-Net was also more efficient than FPS. For example, at $D=2$ it reduced downsampling latency to 0.072 s, compared with 0.144 s for FPS. Overall, CAS-Net maintained detection accuracy while reducing computation relative to FPS, supporting its use in real-time 3D perception.

\begin{figure*}[t]
    \centering

    \begin{subfigure}[b]{0.48\textwidth}
        \centering
        \includegraphics[width=\textwidth]{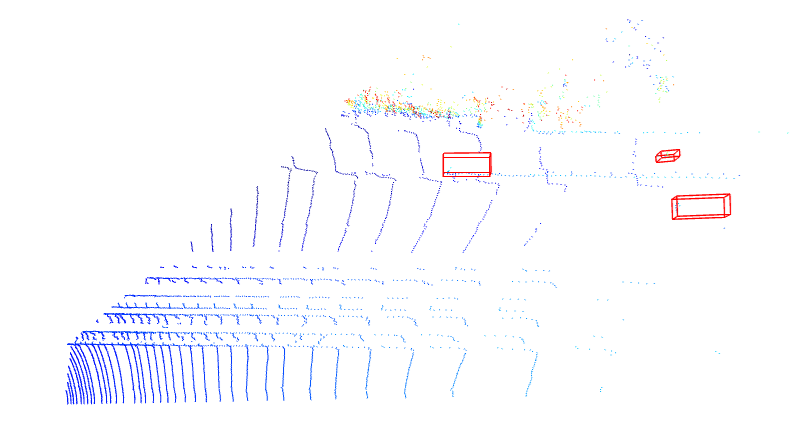}
        \caption{Reference LiDAR Detection (D=1)}
    \end{subfigure}
    \hfill
    \begin{subfigure}[b]{0.48\textwidth}
        \centering
        \includegraphics[width=\textwidth]{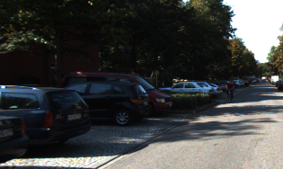}
        \caption{Corresponding RGB Camera Image}
    \end{subfigure}
    
    \vspace{0.5cm} 
    
    \makebox[0.05\textwidth]{}
    \makebox[0.3\textwidth]{\small \textbf{Random Sampling}}
    \makebox[0.3\textwidth]{\small \textbf{Farthest Point Sampling}}
    \makebox[0.3\textwidth]{\small \textbf{CAS-Net (Ours)}} \\
    
    \rotatebox{90}{\makebox[0.15\textwidth]{\small \textbf{D=2}}}
    \begin{subfigure}[b]{0.3\textwidth}
        \centering
        \includegraphics[width=\textwidth]{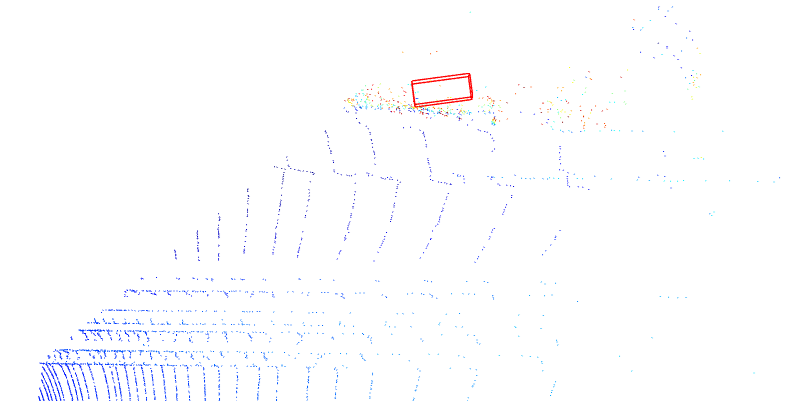}
    \end{subfigure}
    \begin{subfigure}[b]{0.3\textwidth}
        \centering
        \includegraphics[width=\textwidth]{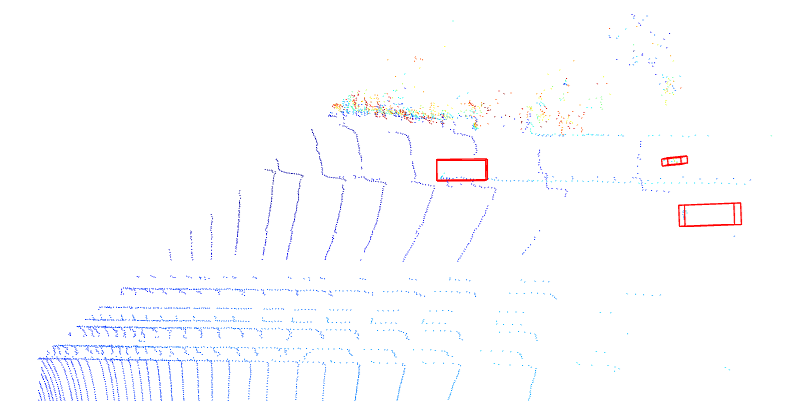}
    \end{subfigure}
    \begin{subfigure}[b]{0.3\textwidth}
        \centering
        \includegraphics[width=\textwidth]{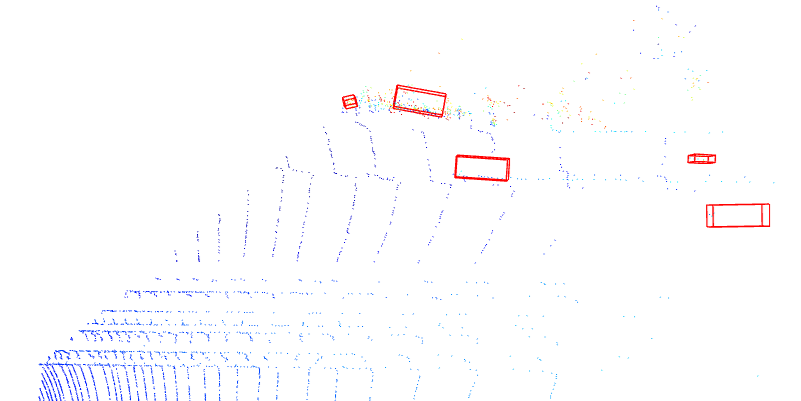}
    \end{subfigure}
    
    \vspace{1mm}
    
    \rotatebox{90}{\makebox[0.15\textwidth]{\small \textbf{D=4}}}
    \begin{subfigure}[b]{0.3\textwidth}
        \centering
        \includegraphics[width=\textwidth]{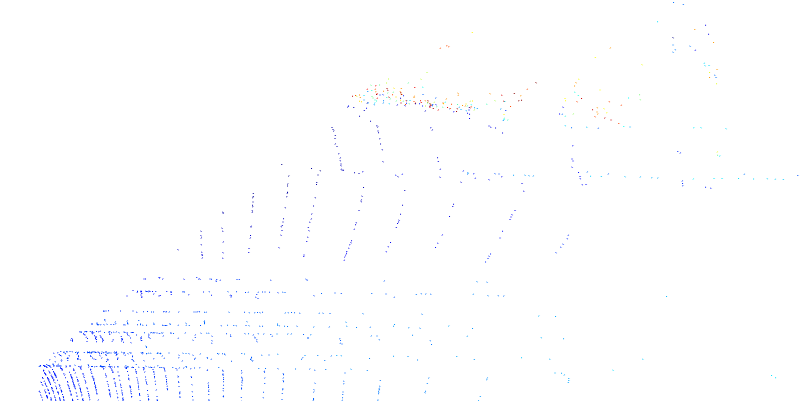}
    \end{subfigure}
    \begin{subfigure}[b]{0.3\textwidth}
        \centering
        \includegraphics[width=\textwidth]{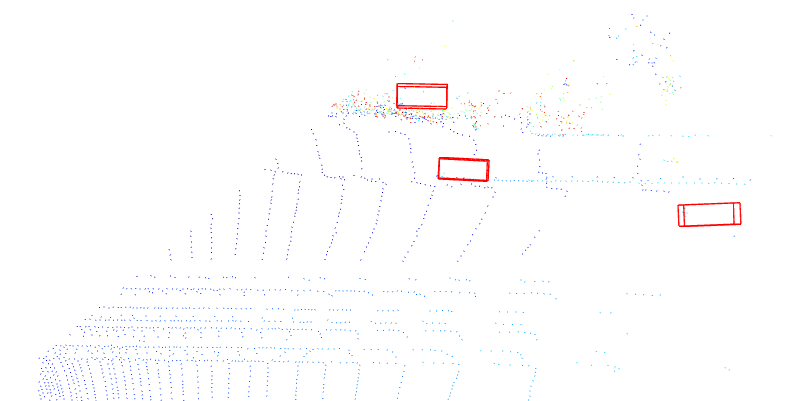}
    \end{subfigure}
    \begin{subfigure}[b]{0.3\textwidth}
        \centering
        \includegraphics[width=\textwidth]{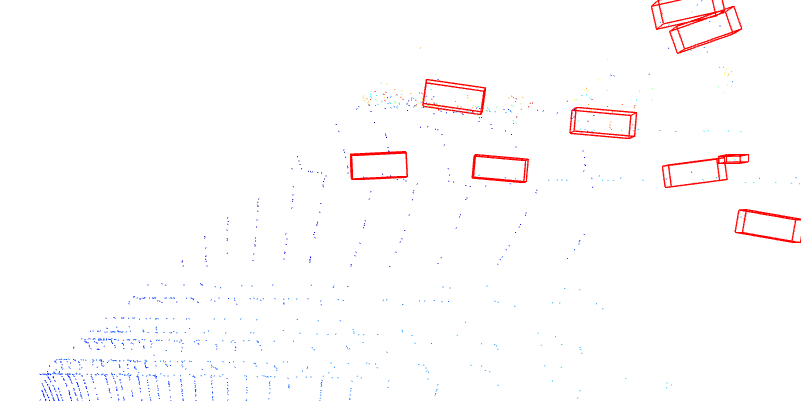}
    \end{subfigure}
    
    \vspace{1mm}

    \rotatebox{90}{\makebox[0.15\textwidth]{\small \textbf{D=8}}}
    \begin{subfigure}[b]{0.3\textwidth}
        \centering
        \includegraphics[width=\textwidth]{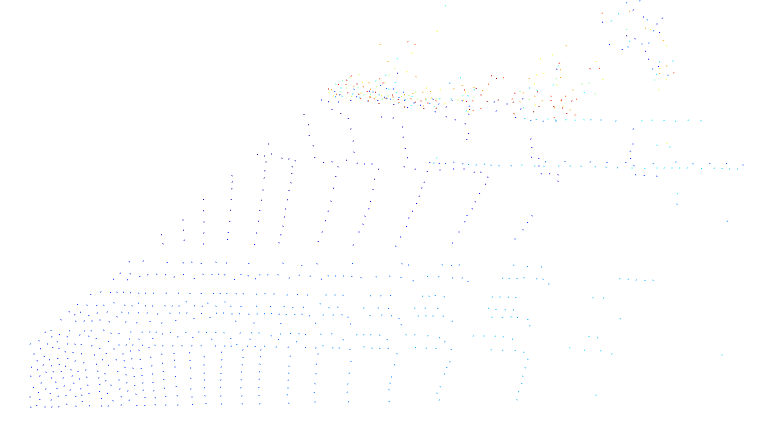}
    \end{subfigure}
    \begin{subfigure}[b]{0.3\textwidth}
        \centering
        \includegraphics[width=\textwidth]{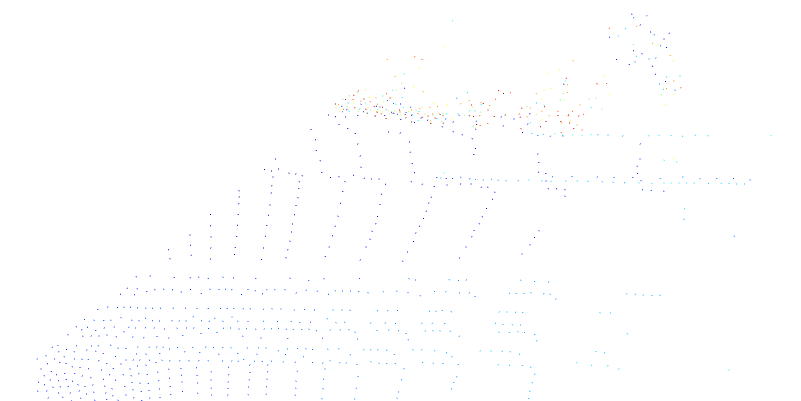}
    \end{subfigure}
    \begin{subfigure}[b]{0.3\textwidth}
        \centering
        \includegraphics[width=\textwidth]{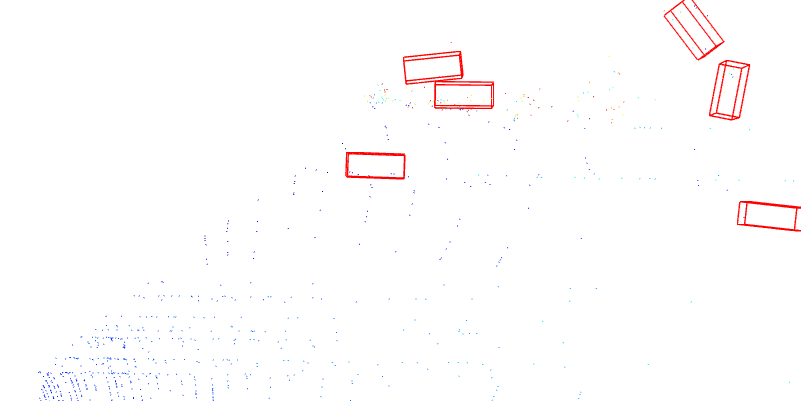}
    \end{subfigure}

    \caption{Qualitative comparison of 3D object detection under different downsampling strategies and ratios. The top row shows the full-resolution reference and the scene context.}
    \label{fig:qualitative_results_det1}
\end{figure*}

Qualitative detection results are shown in Fig.~\ref{fig:qualitative_results_det1}. The figure compares the baseline ($D=1$) with different downsampling strategies. As the downsampling ratio $D$ increased, bounding box stability degraded for both RS and FPS. At $D=8$, RS and FPS failed to preserve sufficient geometric context, which led to many missed detections. In contrast, CAS-Net remained robust: even at high compression, the predicted bounding boxes stayed close to the full-resolution reference.

At high compression ratios, the figure also shows a small increase in false positives. These are mainly low-confidence predictions and can be removed with standard confidence thresholding without reducing detection of the primary objects. Overall, CAS-Net better preserved the structure of salient objects and supported reliable perception when strong data reduction was required.

\subsection{Object classification}
We used PointNet \cite{1} as the backbone network for classification and evaluated CAS-Net AHSN on four datasets: ModelNet40 \footnote{\url{https://github.com/datasets-mila/datasets--modelnet40}}, KITTI \footnote{\url{https://www.cvlibs.net/datasets/kitti/}}, ScanObjectNN \footnote{\url{https://hkust-vgd.github.io/scanobjectnn/}}, and ESTATE \footnote{\url{https://github.com/3DOM-FBK/ESTATE}}. 

ModelNet40 is a synthetic object classification benchmark containing 12,311 CAD models from 40 man-made object categories, split into 9,843 samples for training and 2,468 samples for testing. The number of points in each point cloud was 2048 and reduced to 1024 using FPS. 

For KITTI, the Velodyne LiDAR point clouds from the KITTI Object Detection Benchmark were used as the raw sensing modality. Object-level point clouds were generated by projecting the LiDAR points into the camera coordinate frame using the provided calibration parameters and extracting all points that fall within the annotated three-dimensional bounding boxes in each frame. The bounding boxes and object annotations were obtained from the official label files of the benchmark, which provide the object class, box dimensions, spatial location, and orientation in the rectified camera coordinate system. After transforming the point coordinates into the same reference frame as the annotations, each bounding box was treated as a spatial filter. Points whose coordinates satisfied the geometric constraints of the box were retained to form an individual object point cloud. To ensure sufficient geometric information for downstream learning tasks, only objects containing at least 2048 points were kept. We split the resulting set into training and testing subsets using a 7:3 ratio, while balancing the distribution of point counts across the two subsets. This
produced 1,126 training samples and 485 test samples for
classification. The average number of points in a point cloud
was 6103, ranging from 2048 to 42,592. The number of points
was reduced to 1024 using RS.

ScanObjectNN is a real-world point cloud classification dataset containing 5,000 objects from 15 categories, with 2,902 unique object instances. It provides 2,309 point clouds for training and 581 for testing. Each point cloud contains 2,048 points. The number of points in each point cloud was reduced to 1024 using FPS. 

ESTATE is an object classification dataset constructed by combining object instances extracted from several real-world LiDAR datasets. As only the training subset is currently publicly available, we used this subset for our experiments. Each sample is already provided as an object-level point cloud belonging to one of 13 classes, and we retained only samples containing more than 4096 points. The resulting set was then split into training and evaluation subsets using a 7:3 ratio, while keeping the average number of points per sample similar between the two subsets. This yielded 1,107 training samples and 405 test samples. The average number of points in a point cloud was 14,126, ranging from 4096 to 976,465. The number of points was reduced to 1024 using RS. 
  

CAS-Net was implemented in Pytorch with a batch size of 12, 400 training epochs, and a learning rate of $5 \times 10^{-5}$. We set $\alpha$ = 1, $\beta$ = 1, and \emph{c} = 64. All experiments were conducted on a system with an Intel Core i7-7820X CPU, an NVIDIA GeForce RTX 2080 Ti GPU, and 12 GB memory.

We evaluated the computational cost and classification performance of CAS-Net by comparing the original configuration ($k = 32$, three OA layers) with a reduced configuration ($k = 1$, one OA layer). For all experiments, we report execution time and classification metrics, and we analyze the effect of the sampling ratio (2:1, 4:1, and 8:1) to measure the trade-off between efficiency and performance. 

The study also compares three neighborhood search implementations: PyTorch3D ball query with a radius of 2, PyTorch3D brute-force k-NN (both executed on the GPU), and a k-d tree search using Python’s scipy.spatial (executed on the CPU). All methods were evaluated on the four datasets under both CAS-Net configurations. In the PyTorch3D ball query implementation, if fewer than $k$ neighbors are found within the radius, the remaining indices retain their initialized value (e.g., -1) and are ignored in subsequent computations.

Finally, we compared the fast CAS-Net configuration ($k = 1$, one OA layer) using PyTorch3D ball query with FPS and random sampling. Tables~\ref{tab:modelnet}, \ref{tab:KITTI}, \ref{tab:ScanObjectNN}, and \ref{tab:ESTATE} summarize the results. 

Reducing both the neighborhood size and the number of OA layers consistently decreased computation time across all datasets and search methods. The per-sample execution time was typically reduced by about 41\% to 64\%. The effect on classification performance depended on the dataset. For ModelNet40 and KITTI, the change was generally small, and most configurations showed only minor differences in F1-score, with a few cases showing slight improvement. For ScanObjectNN and ESTATE, the effect was less consistent: the same reduction can improve or reduce performance, with F1-score changes ranging from about +0.03 to -0.07, and recall dropping by up to 3-4 percentage points in some cases.

Increasing the sampling ratio from 2:1 to 4:1 to 8:1 consistently reduced runtime but usually increased performance loss. This trade-off was small on ModelNet40 and KITTI, where performance remained relatively stable across ratios, but it was more visible on ScanObjectNN and ESTATE, where higher sampling ratios more often reduced F1-score and recall.

Comparing the three search methods, ball query provided the most consistent balance between runtime and performance across datasets. Brute-force k-NN often achieved the best classification performance on cleaner datasets (especially ModelNet40), but it was usually the most expensive in computation time. k-d tree was often faster, particularly in several ScanObjectNN settings, but its performance was less consistent.

On ModelNet40, CAS-Net’s performance was comparable to that of FPS, while RS was competitive at moderate sampling ratios but degraded more at higher ratios. In terms of speed, CAS-Net was consistently faster than FPS, while RS was by far the fastest. 

On KITTI, CAS-Net achieved similar classification performance to FPS, while RS performed worse. CAS-Net was much faster than FPS but slower than RS, giving a better speed–accuracy trade-off than FPS and more reliable accuracy than RS. 

On ScanObjectNN, CAS-Net achieved similar classification performance to FPS, while FPS was substantially slower. RS was the fastest method, but it showed a larger performance drop. 

On ESTATE, CAS-Net achieved better classification performance than FPS while running much faster. RS was the fastest method and achieved competitive performance, indicating that this dataset is less sensitive to the sampling strategy.

In summary, CAS-Net was faster than FPS but slower than RS. CAS-Net achieved similar classification performance to FPS, with small gains or losses depending on the setting. RS was the fastest method, but it usually produced lower classification performance than CAS-Net and exceeded the best CAS-Net result in only one setting.

\section{Conclusion}

CAS-Net provided an efficient learned alternative to traditional sampling. It consistently reduced sampling latency relative to FPS while maintaining similar, and in some settings better, performance. The benefits became more apparent as the downsampling ratio increased, where CAS-Net preserved task-relevant structure more reliably than the baselines. RS was the fastest method, but it typically showed a larger and less predictable performance drop and only rarely exceeded the best CAS-Net result. Overall, CAS-Net offered a stable speed–accuracy trade-off that supports its use in resource-constrained perception applications that require substantial data reduction.
Although CAS-Net was faster than FPS, it was still much slower than RS. Reducing the overhead of neighborhood search, for example through approximate nearest neighbours, could improve real-time applicability. The classification experiments also showed that reducing the search range and the number of OA layers improves speed but can change performance in a less predictable way on noisier data. Adaptive settings that adjust the search range, OA depth, or sampling ratio based on input quality or scene complexity may therefore provide more stable accuracy. 

\section{Acknowledgement}

This work was supported in part by the Royal Society International Exchange Scheme IES\textbackslash R2\textbackslash 232053. 

This paper was also supported by the János Bolyai Research Scholarship of the Hungarian Academy of Sciences and projects grant no. STARTING 149552 and K 139485.
Project no. STARTING 149552 and K 139485 were implemented with support provided by the Ministry of Culture and Innovation of Hungary from the National Research, Development and Innovation Fund, financed under the STARTING\_24 and K\_21 funding schemes.

We are grateful for the opportunity to use HUN-REN Cloud (see \cite{Heder2022}; https://science-cloud.hu/), which helped us achieve the results published in this paper.


\bibliographystyle{IEEEbib}
\bibliography{icme2023template}

\begin{table*}[!ht]
	\caption{Performance comparaison of CAS-Net with FPS and RS on ModelNet40. OA denotes the number of OA layers, and $k$ denotes the number of neighboring points. Performance metrics are accuracy (Acc), precision (Prec), recall (Rec), and F1-score.}
	\label{tab:modelnet}
	\centering
	  \begin{tabular*}{0.9\linewidth}{@{\extracolsep{\fill}}@{\hspace{0.5em}}ccccccccccc@{\hspace{0.5em}}@{\extracolsep{\fill}}}
	\toprule
	\multirow{2}{*}{Method} &\multirow{2}{*}{Input points}  &\multirow{2}{*}{Output points} &\multirow{2}{*}{OA} &\multirow{2}{*}{$k$} &\multicolumn{2}{c}{Execution time (s)} &\multirow{2}{*}{Acc} &\multirow{2}{*}{Prec} &\multirow{2}{*}{Rec} &\multirow{2}{*}{F1}\\\cline{6-7}
	& & & & &per batch &per sample & & & \\
	\midrule
	\multirow{6}{*}{CAS-Net Ball query}   
	&1024	&512	&1	&1	&0.019459	&0.001624	&89.51\%	&84.21\%	&85.23\%	&0.8435	\\
	&1024	&256	&1	&1	&0.015566	&0.001299	&89.47\%	&84.92\%	&85.95\%	&0.8511	\\
	&1024	&128	&1	&1	&0.015298	&0.001277	&88.90\%	&85.04\%	&85.31\%	&0.8484	\\
	&1024	&512	&3	&32	&0.049871	&0.004896	&89.38\%	&85.54\%	&85.42\%	&0.8490	\\
	&1024	&256	&3	&32	&0.044244	&0.003693	&89.42\%	&84.80\%	&85.33\%	&0.8478	\\
	&1024	&128	&3	&32	&0.041973	&0.003503	&89.18\%	&84.59\%	&85.08\%	&0.8457	\\\hline
	\multirow{6}{*}{CAS-Net k-NN}   
	&1024	&512	&1	&1	&0.018823	&0.001571	&89.99\%	&84.69\%	&86.16\%	&0.8512	\\
	&1024	&256	&1	&1	&0.015024	&0.001254	&89.34\%	&83.85\%	&86.03\%	&0.8474 \\
	&1024	&128	&1	&1	&0.014254	&0.001190	&88.86\%	&84.39\%	&84.60\%	&0.8401 \\
	&1024	&512	&3	&32	&0.063233	&0.005278	&90.03\%	&85.09\%	&86.62\%	&0.8551 \\
	&1024	&256	&3	&32	&0.058508	&0.004884	&89.59\%	&85.28\%	&84.64\%	&0.8475 \\
	&1024	&128	&3	&32	&0.055698	&0.004649	&89.26\%	&85.36\%	&85.31\%	&0.8510 \\\hline
	\multirow{6}{*}{CAS-Net k-d tree}  
	&1024	&512	&1	&1	&0.019632	&0.001639	&88.37\%	&83.12\%	&83.60\%	&0.8303 \\
	&1024	&256	&1	&1	&0.014497	&0.001210	&88.37\%	&84.19\%	&84.33\%	&0.8391 \\
	&1024	&128	&1	&1	&0.013024	&0.001087	&86.51\%	&81.35\%	&82.39\%	&0.8152 \\
	&1024	&512	&3	&32	&0.047473	&0.003963	&88.25\%	&82.70\%	&84.15\%	&0.8295 \\
	&1024	&256	&3	&32	&0.042081	&0.003446	&87.64\%	&81.64\%	&82.98\%	&0.8167 \\
	&1024	&128	&3	&32	&0.041280	&0.003446	&86.99\%	&81.23\%	&82.44\%	&0.8165 \\\hline
    \multirow{3}{*}{FPS}
    &1024  &512 & & &0.087730 &0.007323 &89.87\% &84.54\% &85.79\% &0.8488 \\
    &1024  &256 & & &0.043298 &0.003614 &89.79\% &85.10\% &84.92\% &0.8479 \\
    &1024  &128 & & &0.022560 &0.001883 &89.14\% &85.41\% &84.65\% &0.8414 \\\hline
    \multirow{3}{*}{RS}
    &1024  &512 & & &0.000540 &4.50E-05 &89.42\% &86.29\% &85.09\% &0.8511 \\
    &1024  &256 & & &0.000517 &4.30E-05 &89.30\% &84.57\% &83.86\% &0.8375 \\
    &1024  &128 & & &0.000519 &4.30E-05 &87.84\% &82.35\% &83.68\% &0.8273 \\\bottomrule
	\end{tabular*}
\end{table*}


\begin{table*}[!ht]
	\caption{Performance comparaison of CAS-Net with FPS and RS on KITTI. OA denotes the number of OA layers, and $k$ denotes the number of neighboring points. Performance metrics are accuracy (Acc), precision (Prec), recall (Rec), and F1-score.}
	\label{tab:KITTI}
	\centering
	\begin{tabular*}{0.9\linewidth}{@{\extracolsep{\fill}}@{\hspace{0.5em}}ccccccccccc@{\hspace{0.5em}}@{\extracolsep{\fill}}}
		\toprule
		\multirow{2}{*}{Method} &\multirow{2}{*}{Input points}  &\multirow{2}{*}{Output points} &\multirow{2}{*}{OA} &\multirow{2}{*}{$k$} &\multicolumn{2}{c}{Execution time (s)} &\multirow{2}{*}{Acc} &\multirow{2}{*}{Prec} &\multirow{2}{*}{Rec} &\multirow{2}{*}{F1}\\\cline{6-7}
		& & & & &per batch &per sample & & & \\
		\midrule
		\multirow{6}{*}{CAS-Net Ball query}   
		&1024	&512	&1	&1	&0.036872	&0.003117	&95.88\%	&95.50\%	&90.45\%	&0.9189 \\
		&1024	&256	&1	&1	&0.030619	&0.002588	&95.88\%	&92.73\%	&90.26\%	&0.9070 \\
		&1024	&128	&1	&1	&0.027265	&0.002305	&94.43\%	&87.64\%	&88.48\%	&0.8777 \\
		&1024	&512	&3	&32	&0.062607	&0.005293	&96.08\%	&93.22\%	&92.17\%	&0.9219 \\
		&1024	&256	&3	&32	&0.058407	&0.004937	&95.05\%	&92.62\%	&91.41\%	&0.9151 \\
		&1024	&128	&3	&32	&0.054242	&0.004585	&94.23\%	&90.00\%	&89.72\%	&0.8830 \\\hline		
		\multirow{6}{*}{CAS-Net k-NN}   
		&1024	&512	&1	&1	&0.039745	&0.003360	&95.26\%	&94.11\%	&89.24\%	&0.9131 \\
		&1024	&256	&1	&1	&0.031431	&0.002657	&95.67\%	&90.60\%	&87.50\%	&0.8813 \\
		&1024	&128	&1	&1	&0.025462	&0.002152	&95.67\%	&92.04\%	&88.58\%	&0.8961 \\
		&1024	&512	&3	&32	&0.074549	&0.006302	&95.88\%	&93.61\%	&90.30\%	&0.9104 \\
		&1024	&256	&3	&32	&0.070138	&0.005929	&96.08\%	&94.42\%	&88.38\%	&0.9020 \\
		&1024	&128	&3	&32	&0.065736	&0.005557	&94.64\%	&89.09\%	&88.75\%	&0.8882 \\\hline		
		\multirow{6}{*}{CAS-Net k-d tree}  
		&1024	&512	&1	&1	&0.034834	&0.002945	&94.43\%	&89.09\%	&90.61\%	&0.8968 \\
		&1024	&256	&1	&1	&0.029360	&0.002482	&95.26\%	&91.58\%	&88.41\%	&0.8920 \\
		&1024	&128	&1	&1	&0.025501	&0.002156	&95.26\%	&90.68\%	&87.92\%	&0.8888 \\
		&1024	&512	&3	&32	&0.059678	&0.005045	&96.08\%	&94.59\%	&90.63\%	&0.9225 \\
		&1024	&256	&3	&32	&0.054813	&0.004634	&95.67\%	&91.78\%	&90.87\%	&0.9105 \\
		&1024	&128	&3	&32	&0.051236	&0.004331	&95.05\%	&92.26\%	&89.37\%	&0.9044 \\\hline
        \multirow{3}{*}{FPS}
        &1024 &512 & & &0.622375 &0.052613 &96.70\% &95.69\% &91.78\% &0.9284 \\
        &1024 &256 & & &0.319921 &0.027045 &95.88\% &92.42\% &91.42\% &0.9157 \\
        &1024 &128 & & &0.159715 &0.013502 &95.67\% &93.66\% &91.37\% &0.9214 \\\hline
        \multirow{3}{*}{RS}
        &1024 &512 & & &0.001078 &9.10E-05 &95.46\% &92.16\% &87.47\% &0.8910 \\
        &1024 &256 & & &0.001014 &8.60E-05 &94.85\% &89.42\% &88.65\% &0.8853 \\
        &1024 &128 & & &0.000986 &8.30E-05 &94.43\% &89.82\% &87.33\% &0.8826 \\\bottomrule
	\end{tabular*}
\end{table*}


\begin{table*}[!ht]
	\caption{Performance comparaison of CAS-Net with FPS and RS on KITTI. OA denotes the number of OA layers, and $k$ denotes the number of neighboring points. Performance metrics are accuracy (Acc), precision (Prec), recall (Rec), and F1-score.}
	\label{tab:ScanObjectNN}
	\centering
	\begin{tabular*}{0.9\linewidth}{@{\extracolsep{\fill}}@{\hspace{0.5em}}ccccccccccc@{\hspace{0.5em}}@{\extracolsep{\fill}}}
		\toprule
		\multirow{2}{*}{Method} &\multirow{2}{*}{Input points}  &\multirow{2}{*}{Output points} &\multirow{2}{*}{OA} &\multirow{2}{*}{$k$} &\multicolumn{2}{c}{Execution time (s)} &\multirow{2}{*}{Acc} &\multirow{2}{*}{Prec} &\multirow{2}{*}{Rec} &\multirow{2}{*}{F1}\\\cline{6-7}
		& & & & &per batch &per sample & & & \\
		\midrule
		\multirow{6}{*}{CAS-Net Ball query}   
		&1024	&512	&1	&1	&0.052503	&0.004428 	&69.71\%	&65.22\%	&62.14\%	&0.6286 \\
		&1024	&256	&1	&1	&0.042908 	&0.003619 	&69.54\%	&68.16\%	&62.64\%	&0.6394 \\
		&1024	&128	&1	&1	&0.040196 	&0.003390 	&68.16\%	&62.97\%	&60.34\%	&0.6056 \\
		&1024	&512	&3	&32	&0.119326 	&0.010064 	&69.54\%	&67.37\%	&60.86\%	&0.6199 \\
		&1024	&256	&3	&32	&0.111203 	&0.009379 	&68.33\%	&65.63\%	&59.82\%	&0.6096 \\
		&1024	&128	&3	&32	&0.109072 	&0.009199 	&69.54\%	&64.65\%	&61.80\%	&0.6223 \\\hline		
		\multirow{6}{*}{CAS-Net k-NN}   
		&1024	&512	&1	&1	&0.050349 	&0.004246 	&68.50\%	&62.50\%	&61.11\%	&0.6061 \\
		&1024	&256	&1	&1	&0.039616 	&0.003341 	&69.02\%	&64.19\%	&60.98\%	&0.6126 \\
		&1024	&128	&1	&1	&0.036453 	&0.003074 	&67.99\%	&63.86\%	&60.27\%	&0.6109 \\
		&1024	&512	&3	&32	&0.175048 	&0.014763	&69.54\%	&65.78\%	&61.76\%	&0.6201 \\
		&1024	&256	&3	&32	&0.169487 	&0.014294 	&68.50\%	&63.48\%	&59.97\%	&0.5994 \\
		&1024	&128	&3	&32	&0.159940 	&0.013489 	&68.16\%	&63.44\%	&61.26\%	&0.6180 \\\hline		
		\multirow{6}{*}{CAS-Net k-d tree}  
		&1024	&512	&1	&1	&0.031208	&0.002632 	&68.67\%	&63.54\%	&60.74\%	&0.6091 \\
		&1024	&256	&1	&1	&0.026963 	&0.002274 	&66.27\%	&61.78\%	&58.06\%	&0.5918 \\
		&1024	&128	&1	&1	&0.025148 	&0.002121 	&61.45\%	&55.32\%	&54.09\%	&0.5383 \\
		&1024	&512	&3	&32	&0.056480 	&0.004763 	&68.50\%	&66.43\%	&61.19\%	&0.6245 \\
		&1024	&256	&3	&32	&0.052221 	&0.004404 	&69.54\%	&66.43\%	&62.18\%	&0.6318 \\
		&1024	&128	&3	&32	&0.050859 	&0.004289 	&67.30\%	&63.13\%	&59.50\%	&0.6014 \\\hline
        \multirow{3}{*}{FPS}
        &1024 &512 & & &0.259297 &0.021868 &69.36\% &66.77\% &63.29\% &0.6402 \\
        &1024 &256 & & &0.134763 &0.011366 &68.33\% &64.41\% &60.48\% &0.6098 \\
        &1024 &128 & & &0.066854 &0.005638 &68.33\% &64.80\% &60.88\% &0.6123 \\\hline
        \multirow{3}{*}{RS}
        &1024 &512 & & &0.000906 &0.000076 &66.61\% &61.57\% &60.29\% &0.6023 \\
        &1024 &256 & & &0.000832 &0.00007  &66.78\% &62.89\% &60.29\% &0.6079 \\
        &1024 &128 & & &0.000814 &0.000069 &65.06\% &59.84\% &56.57\% &0.5729 \\\bottomrule
	\end{tabular*}
\end{table*}


\begin{table*}[!ht]
	\caption{Performance comparison of CAS-Net with FPS and RS on ESTATE. OA denotes the number of OA layers, and $k$ denotes the number of neighboring points. Performance metrics are accuracy (Acc), precision (Prec), recall (Rec), and F1-score.}
	\label{tab:ESTATE}
	\centering
	\begin{tabular*}{0.9\linewidth}{@{\extracolsep{\fill}}@{\hspace{0.5em}}ccccccccccc@{\hspace{0.5em}}@{\extracolsep{\fill}}}
		\toprule
		\multirow{2}{*}{Method} &\multirow{2}{*}{Input points}  &\multirow{2}{*}{Output points} &\multirow{2}{*}{OA} &\multirow{2}{*}{$k$} &\multicolumn{2}{c}{Execution time (s)} &\multirow{2}{*}{Acc} &\multirow{2}{*}{Prec} &\multirow{2}{*}{Rec} &\multirow{2}{*}{F1}\\\cline{6-7}
		& & & & &per batch &per sample & & & \\
		\midrule
		\multirow{6}{*}{CAS-Net Ball query}  
		&1024	&512	&1	&1	&0.036097 	&0.003069 	&84.12\%	&72.70\%	&75.96\%	&0.7292 \\
		&1024	&256	&1	&1	&0.033790 	&0.002873 	&85.01\%	&73.90\%	&72.82\%	&0.7115 \\
		&1024	&128	&1	&1	&0.029574 	&0.002514 	&84.56\%	&72.78\%	&72.94\%	&0.7056 \\
		&1024	&512	&3	&32	&0.062459 	&0.005310 	&85.01\%	&67.70\%	&72.42\%	&0.6764 \\
		&1024	&256	&3	&32	&0.059424 	&0.005052 	&85.68\%	&73.77\%	&76.16\%	&0.7381 \\
		&1024	&128	&3	&32	&0.054903 	&0.004667 	&83.22\%	&66.90\%	&66.50\%	&0.6523 \\\hline
		\multirow{6}{*}{CAS-Net k-NN}   
		&1024	&512	&1	&1	&0.034513 	&0.002934 	&84.56\%	&71.60\%	&72.48\%	&0.6956 \\
		&1024	&256	&1	&1	&0.029149 	&0.002478 	&84.56\%	&67.44\%	&70.33\%	&0.6709 \\
		&1024	&128	&1	&1	&0.026571 	&0.002259 	&83.45\%	&65.14\%	&68.93\%	&0.6505 \\
		&1024	&512	&3	&32	&0.076379 	&0.006493 	&84.12\%	&70.20\%	&72.05\%	&0.6878 \\
		&1024	&256	&3	&32	&0.068949 	&0.005861 	&84.12\%	&64.04\%	&66.16\%	&0.6480 \\
		&1024	&128	&3	&32	&0.069778 	&0.005932 	&85.23\%	&75.06\%	&72.58\%	&0.7178 \\\hline
		\multirow{6}{*}{CAS-Net k-d tree}  
		&1024	&512	&1	&1	&0.036275 	&0.003084 	&85.46\%	&69.29\%	&70.84\%	&0.6874 \\
		&1024	&256	&1	&1	&0.029112 	&0.002475 	&86.80\%	&74.89\%	&75.07\%	&0.7277 \\
		&1024	&128	&1	&1	&0.028231 	&0.002400 	&83.67\%	&67.28\%	&72.23\%	&0.6805 \\
		&1024	&512	&3	&32	&0.062041 	&0.005274 	&84.56\%	&74.53\%	&71.20\%	&0.7123 \\
		&1024	&256	&3	&32	&0.055996 	&0.004760 	&84.12\%	&68.23\%	&71.13\%	&0.6853 \\
		&1024	&128	&3	&32	&0.053789 	&0.004573 	&82.55\%	&65.98\%	&63.99\%	&0.6371 \\\hline
        \multirow{3}{*}{FPS}
        &1024 &512 & & &1.664074 &0.141465 &83.22\% &71.73\% &68.04\% &0.6709 \\
        &1024 &256 & & &0.831077 &0.070651 &82.10\% &65.74\% &69.80\% &0.6488 \\
        &1024 &128 & & &0.406195 &0.034531 &83.45\% &71.89\% &73.84\% &0.7063 \\\hline
        \multirow{3}{*}{RS}
        &1024 &512 & & &0.002280  &0.000194 &86.80\% &75.78\% &78.08\% &0.7555 \\
        &1024 &256 & & &0.002246 &0.000191 &84.34\% &70.22\% &74.64\% &0.7071 \\
        &1024 &128 & & &0.002250  &0.000191 &84.12\% &68.07\% &73.00\% &0.6861 \\\bottomrule
	\end{tabular*}
\end{table*}


\end{document}